\documentclass{article}

\usepackage{arxiv}

\usepackage[utf8]{inputenc} 
\usepackage[T1]{fontenc}    
\usepackage{hyperref}       
\usepackage{url}            
\usepackage{booktabs}       
\usepackage{amsfonts}       
\usepackage{nicefrac}       
\usepackage{microtype}      
\usepackage{lipsum}         
\usepackage{graphicx}
\usepackage{natbib}
\usepackage{doi}

\usepackage{pifont}
\usepackage{nicefrac}
\usepackage{amsthm}
\usepackage{multirow}
\usepackage{subcaption}
\usepackage{color, colortbl}
\usepackage[dvipsnames]{xcolor}
\usepackage{amsmath}

\definecolor{tabhighlight}{HTML}{e5e5e5}
\usepackage{cleveref}       

\newcommand{\tableCellHeight}{1}
\newcommand{\tabstyle}[1]{
	\setlength{\tabcolsep}{#1}
	\renewcommand{\arraystretch}{\tableCellHeight}
	\centering
	\small
}

\newsavebox\CBox
\def\textBF#1{\sbox\CBox{#1}\resizebox{\wd\CBox}{\ht\CBox}{\textbf{#1}}}

\newcommand{\rotbox}[1]{\rotatebox{90}{#1}}

\newcommand{\gray}[1]{\textcolor{gray}{{#1}}} 
\definecolor{tabhighlight}{HTML}{e5e5e5}
\definecolor{citecolor}{HTML}{0071bc}

\title{Compositional Kronecker Context Optimization for Vision-Language Models}

\newif\ifuniqueAffiliation
\uniqueAffiliationtrue

\usepackage{authblk}

\author[1]{%
    Kun Ding\thanks{\texttt{kun.ding@ia.ac.cn}}%
}
\author[1]{%
	Xiaohui Li\thanks{\texttt{xiaohui.li@ia.ac.cn}}%
}
\author[1]{%
	Qiang Yu\thanks{\texttt{qiang.yu@ia.ac.cn}}%
}
\author[1]{%
	Ying Wang\thanks{\texttt{ywang@nlpr.ia.ac.cn}}%
}
\author[1]{%
	Haojian Zhang\thanks{\texttt{zhanghaojian2014@ia.ac.cn}}%
}
\author[1]{%
	Shiming Xiang\thanks{\texttt{smxiang@nlpr.ia.ac.cn}}%
}
\affil{Institute of Automation, Chinese Academy of Sciences, Beijing, China}

\renewcommand{\shorttitle}{\textit{arXiv} Template}

\hypersetup{
pdftitle={A template for the arxiv style},
pdfsubject={q-bio.NC, q-bio.QM},
pdfauthor={David S.~Hippocampus, Elias D.~Striatum},
pdfkeywords={Vision-Language Models, Prompt Tuning, Context Optimization, Image Recognition},
}

\begin{document}
\maketitle

\begin{abstract}
Context Optimization (CoOp) has emerged as a simple yet effective technique for adapting CLIP-like vision-language models to downstream image recognition tasks. Nevertheless, learning compact context with satisfactory base-to-new, domain and cross-task generalization ability while adapting to new tasks is still a challenge. To tackle such a challenge, we propose a lightweight yet generalizable approach termed Compositional Kronecker Context Optimization (CK-CoOp). Technically, the prompt's context words in CK-CoOp are learnable vectors, which are crafted by linearly combining base vectors sourced from a dictionary. These base vectors consist of a non-learnable component obtained by quantizing the weights in the token embedding layer, and a learnable component constructed by applying Kronecker product on several learnable tiny matrices. Intuitively, the compositional structure mitigates the risk of overfitting on training data by remembering more pre-trained knowledge. Meantime, the Kronecker product breaks the non-learnable restrictions of the dictionary, thereby enhancing representation ability with minimal additional parameters. Extensive experiments confirm that CK-CoOp achieves state-of-the-art performance under base-to-new, domain and cross-task generalization evaluation, but also has the metrics of fewer learnable parameters and efficient training and inference speed.
\end{abstract}


\section{Introduction}
\label{sec:intro}
Prompt tuning~\cite{PrefixTuning,ScalePower,FactualProbe} has garnered extensive research in the field of natural language processing (NLP), serving as a pivotal approach for adapting large language models to various downstream tasks. Recently, the concept of soft prompt tuning~\cite{CoOp,cocoop} has been investigated as a means to adapt vision-language models (VLMs)~\cite{CLIP,FILIP,ALIGN}. In order to adapt contrastive vision-language models, such as CLIP~\cite{CLIP}, this approach involves inserting some learnable vectors into the input text sequence and subsequently optimizing them on downstream datasets. These newly introduced vectors are referred as context, and the corresponding technique is known as  Context Optimization (CoOp)~\cite{CoOp}.

CoOp has achieved great success in image recognition. Nevertheless, a significant challenge persists: its poor generalization ability~\cite{cocoop,ProGrad}. In open-world applications, we expect the tuned models can not only perform well on seen tasks, but also retain their zero-shot learning ability for unforeseen tasks~\cite{cocoop}. We conjecture that the poor generalization ability of CoOp stems from learning the context from scratch without enforcing any structural constraints, leading to a tendency to overfit on limited training data. Intuitively, as the vocabulary in the token embedding layer of pre-trained models is quite large, it is reasonable to assume that the vector space spanned by the context is a subspace of the space spanned by the token embedding dictionary of pre-trained model. 

Based on the above assumption, we select some embedding vectors in the dictionary to construct the context. As such, the context can better maintain the pre-trained knowledge, which not only makes it more data-efficient, but also reduces the risk of overfitting. However, the selection process has computational difficulty. We instead approximate it with a linear combination on some cluster centers of the dictionary. The centers are non-learnable while the combination coefficients are learnable. To further boost the representation ability of this compositional form, we add a learnable bias to the compressed dictionary consisted of cluster centers. To avoid incurring too many extra parameters, we restrict the bias to have a Kronecker form. As such, only some tiny matrices are introduced. We name the proposed context optimization method as Compositional Kronecker Context Optimization (CK-CoOp) in consideration of its special structure.

We have conducted extensive experiments on 15 recognition datasets to validate the effectiveness of CK-CoOp. The results show that CK-CoOp outperforms ProGrad significantly in base-to-new and cross-task generalization, while still achieving comparable performance in domain generalization. Impressively, CK-CoOp manages to accomplish this with a mere 38\% of the parameters and up to 75\% of the training time required by ProGrad, demonstrating its efficiency and effectiveness. Besides, it not only surpasses CoCoOp in base-to-new generalization evaluation but also matches its domain and cross-task generalization performance, while boasting a remarkable reduction of 91\% in parameters, 80\% less training time, and an astonishing 100$\times$ faster inference speed.

In summary, we introduce Compositional Kronecker Context Optimization (CK-CoOp) as a novel approach for tuning Vision-Language Models (VLMs). It enhances generalization ability by crafting a lightweight context informed by unique structural constraints. This innovative structured design distinguishes our method from traditional prompt tuning techniques. In Section~\ref{sec:related_work}, we provide a comprehensive overview of related works. Section~\ref{sec:method} delves into the principles of our proposed method, while Section~\ref{sec:exp} presents and analyzes various experimental results.

\section{Related Work}
\label{sec:related_work}
\textbf{Vision-Language Models (VLMs)} have revolutionized our ability to bridge the gap between vision and language modalities in the latent feature space. By leveraging the vast amount of paired image and text data available on the web, these models have achieved remarkable progress. Broadly speaking, VLMs can be categorized into two major groups: generative and discriminative. Generative VLMs~\cite{WritePaint,VLBeit} typically employ techniques such as image captioning, masked language modeling (MLM), masked image modeling (MIM) for effective representation learning. In contrast, discriminative VLMs~\cite{CLIP,ALIGN,UniCL,FILIP,Florence} frequently adopt cross-modal contrastive learning technique. Notably, VLMs like CLIP~\cite{CLIP} have found extensive applications in various tasks, encompassing image classification~\cite{CoOp,cocoop}, object detection~\cite{DenseCLIP} and image segmentation~\cite{CRIS}. In this work, we leverage the widely-used CLIP model as our foundation to investigate approaches for enhancing the generalization capabilities of CoOp. Nevertheless, it is worth noting that other contrastive VLMs could also be explored and potentially applied in our setting.

\textbf{Prompt Tuning} is one of the classical prompt learning methods, which originates from the NLP domain~\cite{PrefixTuning,ScalePower,FactualProbe}. Distinguishing itself from earlier techniques such as prompt designing~\cite{BrownMRSKDNSSAA20,RaffelSRLNMZLL20} and prompt searching~\cite{GaoFC20,ShinRLWS20}, prompt tuning focuses on optimizing continuous vectors in the word embedding space. This approach effectively circumvents the laborious and time-consuming task of manual design, albeit at the expense of some interpretability. \cite{PromptSurvey} provides a comprehensive survey of prompt learning in NLP.

\textbf{Context Optimization (CoOp)}~\cite{CoOp} demonstrates a successful use case of prompt tuning to VLMs for the image recognition task. It prepends learnable and continuous vectors as a context to various class texts and optimizes the context with softmax classification loss. CoOp significantly boosts few-shot recognition performance on many downstream tasks. Despite its success, it exhibits unsatisfactory generalization ability. In view of this, CoCoOp~\cite{cocoop} proposed instance-conditioned context optimization, employing a meta network to generate context based on global image features. This approach significantly improves base-to-new and domain generalization abilities but comes at the cost of lowering accuracy on base classes and training and inference efficiency. ProGrad~\cite{ProGrad} handled this problem by updating the context whose gradient is aligned to the general direction. Unlike these methods, we tackle the generalization challenge by designing a structural context that is compact and effectively reduces the parameter count, training and inference time. Although there are other related works~\cite{DualCoOp,SoftCPT,PDL,NPS}, they do not primarily focus on resolving the generalization problem.
\begin{figure*}[t!]
	\centering
	\includegraphics[width=0.98\linewidth]{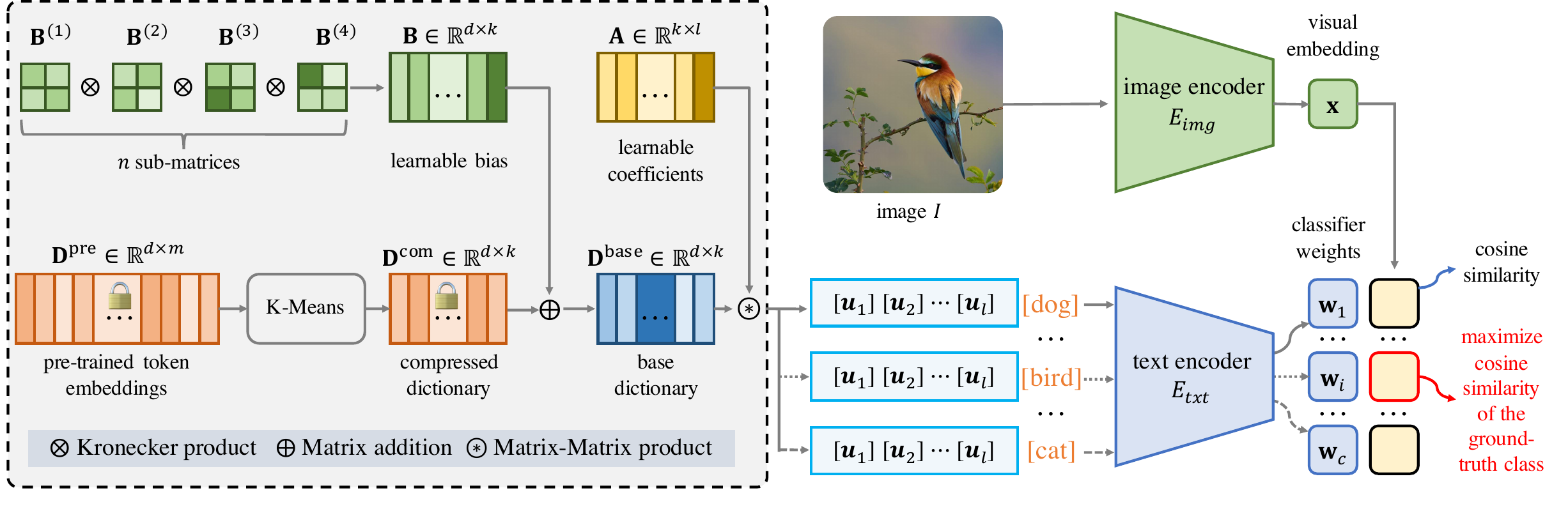}
	\caption{Framework of the proposed CK-CoOp. Different from CoOp~\cite{CoOp}, CK-CoOp imposes a compositional constraint on the context. The left part in the dashed box corresponds to the process of generating the constrained context proposed in this work. Removing this part results in the original CoOp.}
	\label{fig:framework}
	\vspace{0pt}
\end{figure*}

\textbf{Structured Matrix} is broadly used in computer vision and machine learning for regularization~\cite{SR,XiangNMPZ12}, compression~\cite{YuLWT17,ZhangZHS16,CirCNN,YinSL021,0022KDSG17} and speedup~\cite{0022KDSG17,ZhangYGKWC15,YuKGC14}. For example, \cite{XiangNMPZ12} exploited the $L_{2,1}$ group sparsity regularization for feature selection. \cite{YuLWT17} employed low rank and sparse decomposition to compress deep neural networks. \cite{0022KDSG17} proposed a $L_1$ norm based method for filtering useless filters in deep convolutional neural networks. \cite{CircConv} imposed the circulant structure on convolutional layers, leading to reduced complexity. \cite{ZhangYGKWC15} employed the orthogonal Kronecker product for fast binary embedding and quantization. Different from these works, our work exploits the idea of structured matrix designing in context optimization. The bias matrix of the dictionary in this work also has a Kronecker form, but we enforce each sub-matrix to be $L_1$ normalized, which is different from~\cite{ZhangYGKWC15}, and we use it in a different filed. To the best of our knowledge, the concept of structuring the context in prompt tuning has been scarcely explored in previous works. While there may be some works related to our approach, the most pertinent one is \cite{SPT}. However, the context in \cite{SPT} is solely conditioned on task embedding, lacking the structural constraints that we have imposed in our method.

\section{Method}
\label{sec:method}
This part first gives a brief overview of context optimization in Sec.~\ref{ssec:pre}, and then presents the key ingredients of CK-CoOp in Sec.~\ref{ssec:ckcoop}.

\subsection{Preliminary}
\label{ssec:pre}
Context Optimization (CoOp)~\cite{CoOp} introduces prompt tuning, first proposed in NLP, to the adaption of VLMs on downstream image recognition task. It defines a learnable context consisted of $l$ vectors, which is denoted as $\mathbf{U}=[\mathbf{u}_1,\cdots,\mathbf{u}_l]$. The context is prepended to the embedding sequence of each class name, denoted as $\mathbf{C}_i$, where $i$ denotes the $i$-th class.  CoOp generates the classifier weight for the $i$-th class by employing the text encoder $E_{txt}$ in CLIP, i.e., $\mathbf{w}_i=E_{txt}([\mathbf{U}, \mathbf{C}_i])$\footnote{The last layer's embedding of [EOT] token is used as the global text feature.}. Given an image $I$, CoOp extracts its visual embedding $\mathbf{x}$ by CLIP's image encoder $E_{img}$, i.e., $\mathbf{x}=E_{img}(I)$. To classify the image $I$, posterior probability of class $y$ is computed as
\begin{equation}
p(y|\mathbf{x})=\frac{\exp(\cos(\mathbf{x}, \mathbf{w}_y)/\tau)}{\sum\nolimits_{i=1}^c \exp(\cos(\mathbf{x}, \mathbf{w}_i)/\tau)},
\end{equation}
where $\cos(\cdot)$ is the cosine similarity and $\tau$ is the temperature parameter. CoOp keeps all parameters except the context $\mathbf{U}$ fixed and optimizes it by minimizing the negative log likelihood loss on a training set.

\subsection{CK-CoOp}
\label{ssec:ckcoop}
Fig.~\ref{fig:framework} gives the computational flowchart of the proposed CK-CoOp. The cornerstone of CK-CoOp lies in structuring of the context, distinguishing it from CoOp. We design a base dictionary, which is a compressed and tuned version of the original token embedding dictionary in the text encoder of pre-trained VLMs. The compression is achieved by K-Means clustering, while the tuning is realized by attaching a learnable bias which has a compact Kronecker form. Similar to sparse representation and reconstruction~\cite{SR}, the vectors in the context are generated by combining the columns of the base dictionary in a compositional manner. Following parts will detail the above process formally.

\textbf{Dictionary Based Compositional Context.} As previously mentioned, the token embeddings of word pieces in pre-trained VLMs constitute a comprehensive concept pool for words. It is reasonable to assume that the learned context vectors reside within the space defined by these embeddings. By combing some of the embeddings, we can generate the required vectors for the context. Nevertheless, directly employing the entire space of all token embeddings is computationally expensive. Consequently, we opt to compress this space through vector quantization. Let us arrange all $m$ token embeddings with each having a dimension of $d$ in a matrix termed $\mathbf{D}^{\text{pre}}\in \mathbb{R}^{d\times m}$ column by column. By applying K-Means algorithm to cluster the columns into $k$ centers, a compressed dictionary denoted as $\mathbf{D}^{\text{com}}\in \mathbb{R}^{d\times k}$ comprising of these centers is obtained. For convenience, we first omit the bias matrix. Thus, $\mathbf{D}^{\text{com}}\in \mathbb{R}^{d\times k}$ forms the base dictionary $\mathbf{D}^{\text{base}}\in \mathbb{R}^{d\times k}$, directly. For each vector $\mathbf{u}$ (omitting the index) in the context, a learnable coefficient vector $\mathbf{a}\in \mathbb{R}^k$ is introduced. Thereby, $\mathbf{u}$ with the compositional form can be expressed as
\begin{equation}
\mathbf{u}=\mathbf{D}^{\text{base}}\cdot\mathbf{a}.
\end{equation}
The dictionary size $k$ acts the role of balancing the representation ability and data-efficiency. As $d$ may vary for different pre-trained models, we introduce a new parameter $\alpha$ to control $k$ in a relative manner, that is $k=\text{round}(\alpha\cdot d)$ with $\text{round}(\cdot)$ the rounding function.

The designed compositional form has clear relation to sparse representation and reconstruction. Actually, $\mathbf{u}$ can be approximated by
\begin{equation}
\mathbf{u}\approx \mathbf{D}^{\text{pre}}\cdot \tilde{\mathbf{a}},\; \tilde{\mathbf{a}}=\mathbf{E}\cdot\mathbf{a}, 
\end{equation}
where $\mathbf{E}\in \mathbb{R}^{m\times k}$ is a 0-1 matrix, whose $i,j$-th value is 1 if the $i$-th column of $\mathbf{D}^{\text{pre}}$ is the nearest neighbor of the $j$-th column of $\mathbf{D}^{\text{base}}$. The number of non-zero elements in $\tilde{\mathbf{a}}$ is up to $k$. Therefore, $\tilde{\mathbf{a}}$ is a sparse vector given $k\ll m$. Note that, when the K-Centroids clustering is used, the above approximation is accurate.

\textbf{Kronecker Product Based Bias Matrix.} The base dictionary is fixed and this will limit the representation capability of the context. To mitigate this problem, we introduce a learnable bias $\mathbf{B}\in\mathbb{R}^{d\times k}$ to base dictionary, i.e.,
\begin{equation}
\mathbf{D}^{\text{base}}=\mathbf{D}^{\text{com}}+\mathbf{B}.
\end{equation}
To avoid increasing too many extra parameters, the bias matrix is enforced to have a Kronecker form. To be more general, we extract the first $d$ rows and $k$ columns of a larger bias matrix $\tilde{\mathbf{B}}\in\mathbb{R}^{\tilde{d}\times \tilde{k}}$ to construct $\mathbf{B}$, i.e., $\mathbf{B}=\tilde{\mathbf{B}}_{1:d, 1:k}$. 
Meantime, $\tilde{\mathbf{B}}$ is defined by the Kronecker product of $p$ sub-matrices:
\begin{align}
\tilde{\mathbf{B}}&=\text{Norm}(\mathbf{B}^{(1)})\otimes \cdots \otimes\text{Norm}(\mathbf{B}^{(i)})\otimes \cdots \otimes \text{Norm}(\mathbf{B}^{(p)}),\nonumber \\
&=\mathop{\otimes}\nolimits_{i=1}^p \text{Norm}(\mathbf{B}^{(i)}), 
\end{align}
where $\text{Norm}(\cdot)$ is a normalization function, and $\mathbf{B}^{(i)}\in \mathbb{R}^{d_i\times k_i}$ is the $i$-th sub-matrix. The dimensions of the sub-matrices should meet the conditions: $\prod\nolimits_{i=1}^{p}d_i =\tilde{d}$ and $\prod\nolimits_{i=1}^{p}k_i =\tilde{k}$. The normalization is to make the numerical optimization easier. The symbol $\otimes$ denotes the Kronecker product between two matrices. For example, the Kronecker product between $\mathbf{B}^{(1)}$ and $\mathbf{B}^{(2)}$ is defined as
\begin{equation}
\mathbf{B}^{(1)}\otimes \mathbf{B}^{(2)}=
\begin{bmatrix}
B^{(1)}_{1,1}\cdot\mathbf{B}^{(2)} & \dots  & B^{(1)}_{1,k_1}\cdot\mathbf{B}^{(2)} \\
B^{(1)}_{2,1}\cdot\mathbf{B}^{(2)} & \dots  & B^{(1)}_{2,k_1}\cdot\mathbf{B}^{(2)} \\
\vdots & \ddots & \vdots \\
B^{(1)}_{d_1,1}\cdot\mathbf{B}^{(2)} & \dots  & B^{(1)}_{d_1,k_1}\cdot\mathbf{B}^{(2)}
\end{bmatrix},
\end{equation}
where $B^{(1)}_{i,j}$ is the $i,j$-th element of $\mathbf{B}^{(1)}$.

Given the expected number of rows and columns $s$ of each sub-matrix, the actual number\footnote{The gap between the expected and actual number is caused by the non-square form of base dictionary.} of rows and columns of the $i$-th sub-matrix are calculated respectively by
\begin{equation}
d_i=\begin{cases}
s & \!i \leq n_1 \\
\lceil d/\prod\nolimits_{i=1}^{n_1} d_i \rceil & \!i=n_1\!+\!1 \\
1 & \! n_1\!+\!1<i\leq \max(n_1, n_2)\!+\!1
\end{cases}
\end{equation}
and
\begin{equation}
k_i=\begin{cases}
s & \!i \leq n_2 \\
\lceil k/\prod\nolimits_{i=1}^{n_2} k_i \rceil & \!i=n_2\!+\!1 \\
1 & \! n_2\!+\!1<i\leq \max(n_1, n_2)\!+\!1
\end{cases}
\end{equation}
where $n_1=\lfloor \log_s d\rfloor, n_2=\lfloor \log_s k\rfloor$. $\lfloor\cdot\rfloor$ and $\lceil\cdot\rceil$ denote the floor and ceiling function, respectively. Smaller $s$ means using more but smaller sub-matrices, while larger $s$ means using fewer but larger sub-matrices.

\textbf{Parameter Count.} To demonstrate CK-CoOp can reduce the learnable parameter count, we here compare the parameter count of CoOp and CK-CoOp. For CoOp, the shared context needs to be optimized, the parameter count is $dl$. For CK-CoOp, the base dictionary has about $as^2+bs+1$ learnable parameters, where $a=\log_s(\min(d,\alpha d))$ and $b=\log_s(\max(d,\alpha d)) - \log_s(\min(d,\alpha d))$\footnote{Here, we assume that $d$ and $\alpha d$ are powers of $s$. }. Therefore, CK-CoOp has $\alpha dl+as^2+bs+1$ parameters in total. To reduce parameter count, $\alpha$ should be less than 1. For example, when $\alpha=0.375, d=512, l=16, s=2$, the percentage of reduced parameters of CK-CoOp compared to CoOp is 62\%.

\section{Experiments}
\label{sec:exp}
In Sec.~\ref{ssec:setup}, we detail the experiment setup. After that, similar to prior works~\cite{CoOp,cocoop,ProGrad}, we evaluate CK-CoOp under three commonly adopted settings: base-to-new generalization (in Sec.~\ref{ssec:b2n}), domain generalization (in Sec.~\ref{ssec:dg}) and cross-task generalization (in Sec.~\ref{ssec:cdt}). To demonstrate that CK-CoOp is a lightweight yet effective method, we further compare it with state-of-the-art methods across five dimensions in Sec.~\ref{ssec:muld_compare}: learnable parameter count, training time, inference time, overall base-to-new generalization accuracy, overall domain generalization accuracy and overall cross-task generalization accuracy. Finally, in Sec.~\ref{ssec:ab_study}, we conduct an ablation study to gain a deeper understanding of the contributions of various components within CK-CoOp.

\subsection{Setup}
\label{ssec:setup}
\textbf{Datasets.} Following prior works~\cite{CoOp,cocoop,ProGrad}, for the base-to-new and cross-task generalization settings, we use 11 image recognition datasets: ImageNet~\cite{ImageNet} and Caltech101~\cite{Caltech101} for generic object classification; Food101~\cite{Food101}, OxfordPets~\cite{OxfordPets}, StanfordCars~\cite{StanfordCars},  Flowers102~\cite{Flowers102} and FGVCAircraft~\cite{FGVCAircraft} for fine-grained image classification; UCF101~\cite{UCF101} for action recognition; EuroSAT~\cite{EuroSAT} for satellite image classification; SUN397~\cite{SUN397} for scene recognition; DTD~\cite{DTD} for texture classification. For the domain generalization setting, we use ImageNet as the source domain and other four datasets, i.e. ImageNetV2~\cite{ImageNetV2}, ImageNet-Sketch~\cite{ImageNetS}, ImageNet-A~\cite{ImageNetA} and ImageNet-R~\cite{ImageNetR} as the target domains.

\textbf{Compared Methods.} Except for comparing with CoOp, we also compare CK-CoOp with more advanced methods that focusing on addressing generalization problem, including CoCoOp~\cite{cocoop} and ProGrad~\cite{ProGrad}. Meanwhile, the results of zero-shot CLIP~\cite{CLIP} are also reported.

\textbf{Implementation Details.} For CoOp, the context length is set to be 16 and random initialization is adopted. The batch size is 32 and 50 epochs are trained. For CoCoOp, the context is initialized by ``a photo of a''. The batch size is set as 1 and 10 epochs are trained. The training setting of ProGrad is identical to CoOp. As for CK-CoOp, the coefficient matrix and all sub-matrices are initialized by a normal distribution with the mean of zero and standard deviation of 0.02. Identical to CoOp, the context length is fixed as 16. Unless otherwise stated, the parameters $\alpha$ and $s$ are set to be 0.375 and 2, respectively. Besides, all methods use the same learning rate scheduler and optimizer. In the ablation study, ViT-B/16 is adopted as the vision backbone.

To ensure reliable model evaluation, we conduct three experiments with different random seeds and report the average score across all trials. Consistent with prior research, we employ per-class accuracy as the evaluation metric for Caltech101, FGVCAircraft, Flowers102 and Oxford-Pets datasets. For the remaining datasets, we utilize top-1 accuracy as the evaluation criterion.

\begin{table}[t]
	\tabstyle{19pt}
	\caption{Overall results of base-to-new generalization with different vision backbones for 16 shots.}
	\label{tab:b2n_overall}
	\renewcommand{\arraystretch}{1.0}
	\begin{tabular}{l |c |c c r |c}
		\toprule
		&  Backbone & Base & New & Gap & H$\uparrow$ \\
		\midrule
		\gray{CLIP~\cite{CLIP}} & \gray{ResNet-101} & \gray{64.07}& \gray{69.42} & \gray{5.35}&\gray{66.64}\\
		CoOp~\cite{CoOp}& ResNet-101& {79.25}& 61.45 & {17.80}& 69.22\\
		CoCoOp~\cite{cocoop}& ResNet-101 & 76.76 & 65.81 & {10.95} & 70.86\\
		ProGrad~\cite{ProGrad} & ResNet-101 & 78.41 & 65.81 & {12.60} & \underline{71.56} \\
		\rowcolor{tabhighlight}
		CK-CoOp & ResNet-101 & 77.93 & {67.53} & {10.40} & \textBF{72.36}\\
		\midrule
		\gray{CLIP~\cite{CLIP}} & \gray{ViT-B/32}& \gray{67.28} & \gray{71.41} &\gray{4.13}&\gray{69.28}\\
		CoOp~\cite{CoOp} & ViT-B/32& {79.24} & 62.77 & {16.49} & 70.05 \\
		CoCoOp~\cite{cocoop} & ViT-B/32 & {76.26}& 66.63&{9.63}& 71.12 \\
		ProGrad~\cite{ProGrad} & ViT-B/32 & {78.75}& 66.19& {12.56} &\underline{71.93} \\
		\rowcolor{tabhighlight}
		CK-CoOp & ViT-B/32 & 77.86 & {68.26} & {9.60} & \textBF{72.74}\\
		\midrule
		\gray{CLIP~\cite{CLIP}} & \gray{ViT-B/16} & \gray{69.16} & \gray{74.13} & \gray{4.97} & \gray{71.56} \\
		CoOp~\cite{CoOp} & ViT-B/16 &  {82.29} & 67.63 & {14.66} & 74.24 \\
		CoCoOp~\cite{cocoop} & ViT-B/16 &  80.12 & 72.36 &{7.76}& \underline{76.04} \\
		ProGrad~\cite{ProGrad} & ViT-B/16 &  82.27 & 70.10 &{12.17}& 75.70\\
		\rowcolor{tabhighlight}
		CK-CoOp & ViT-B/16 &  80.97 & 73.64 & {7.33} & \textBF{77.13} \\
		\bottomrule
	\end{tabular}
	\vspace{-7pt}
\end{table}

\begin{table*}[!t]
	\tabstyle{3pt}
	\caption{Comparison of different methods in the base-to-new generalization setting using 16 shots. H: Harmonic mean.}
	\label{tab:base_to_novel}
	\begin{subtable}[t]{.3\textwidth}
		\centering
		\caption{ImageNet.}
		\setlength\tabcolsep{2.5pt}
		\renewcommand{\arraystretch}{1.0}
		\begin{tabular}{l cc|c}
			\toprule
			& Base & Novel & H \\
			\midrule
			\gray{CLIP} & \gray{72.38} & \gray{68.10} & \gray{70.17}\\
			\midrule
			CoOp & {76.46} & 65.47 & 70.54\\
			CoCoOp & 75.86 & \textBF{70.59} & \textBF{73.13} \\
			ProGrad & \textBF{76.73}& 67.64& 71.90\\
			\rowcolor{tabhighlight}
			CK-CoOp & 76.47  & 69.55 & 72.85\\
			\bottomrule
		\end{tabular}
	\end{subtable}
	\vspace{0.1em}
	\hspace{\fill}
	\begin{subtable}[t]{.3\textwidth}
		\centering
		\caption{Caltech101.}
		\setlength\tabcolsep{2.5pt}
		\renewcommand{\arraystretch}{1.0}
		\begin{tabular}{l cc|c}
			\toprule
			& Base & Novel & H \\
			\midrule
			\gray{CLIP} & \gray{94.58} & \gray{94.10} & \gray{94.34} \\
			\midrule
			CoOp & {95.74} & 93.65 & 94.68 \\
			CoCoOp & 95.85 & 94.45 & {95.14} \\
			ProGrad & \textBF{96.41} & 93.69& 95.03\\
			\rowcolor{tabhighlight}
			CK-CoOp & 96.11 & \textBF{94.68} & \textBF{95.39}\\
			\bottomrule
		\end{tabular}
	\end{subtable}
	\vspace{0.1em}
	\hspace{\fill}
	\begin{subtable}[t]{.3\textwidth}
		\centering
		\caption{OxfordPets.}
		\setlength\tabcolsep{2.5pt}
		\renewcommand{\arraystretch}{1.0}
		\begin{tabular}{l cc|c}
			\toprule
			& Base & Novel & H \\
			\midrule
			\gray{CLIP} & \gray{91.05} & \gray{97.16} & \gray{94.01} \\\midrule
			CoOp & 94.99 & 93.06 & 94.02 \\
			CoCoOp & {95.00} & \textBF{97.81} & \textBF{96.38} \\
			ProGrad & \textBF{95.44}& 95.15& 95.29\\
			\rowcolor{tabhighlight}
			CK-CoOp & 95.21 & 97.41 & 96.30\\
			\bottomrule
		\end{tabular}
	\end{subtable}
	\\
	\begin{subtable}[t]{.3\textwidth}
		\centering
		\caption{StanfordCars.}
		\setlength\tabcolsep{2.5pt}
		\renewcommand{\arraystretch}{1.0}
		\begin{tabular}{l cc|c}
			\toprule
			& Base & Novel & H \\
			\midrule
			\gray{CLIP} & \gray{63.74}& \gray{74.89} & \gray{68.87} \\
			\midrule
			CoOp & \textBF{78.37} & 67.49 & 72.52 \\
			CoCoOp & 70.73 & 72.70 & {71.70} \\
			ProGrad & 76.91& 70.55& \textBF{73.59}\\
			\rowcolor{tabhighlight}
			CK-CoOp & 72.44 & \textBF{72.67} & 72.55 \\
			\bottomrule
		\end{tabular}
	\end{subtable}
	\vspace{0.1em}
	\hspace{\fill}
	\begin{subtable}[t]{.3\textwidth}
		\centering
		\caption{Flowers102.}
		\setlength\tabcolsep{2.5pt}
		\renewcommand{\arraystretch}{1.0}
		\begin{tabular}{l cc|c}
			\toprule
			& Base & Novel & H \\
			\midrule
			\gray{CLIP} & \gray{70.24} & \gray{77.98} & \gray{73.91} \\\midrule
			CoOp & \textBF{97.23} & 62.40 & 76.02 \\
			CoCoOp & 94.79 & 68.44 & {79.49} \\
			ProGrad & 96.43& 69.10& 80.51\\
			\rowcolor{tabhighlight}
			CK-CoOp & 95.56 & \textBF{70.97} & \textBF{81.45}\\
			\bottomrule
		\end{tabular}
	\end{subtable}
	\vspace{0.1em}
	\hspace{\fill}
	\begin{subtable}[t]{.3\textwidth}
		\centering
		\caption{Food101.}
		\setlength\tabcolsep{2.5pt}
		\renewcommand{\arraystretch}{1.0}
		\begin{tabular}{l cc|c}
			\toprule
			& Base & Novel & H \\
			\midrule
			\gray{CLIP} & \gray{89.77} & \gray{91.29} & \gray{90.52} \\\midrule
			CoOp & 89.54 & 90.14 & 89.84 \\
			CoCoOp & \textBF{90.55} & {91.38} & {90.96} \\
			ProGrad & 90.46& 90.54& 90.50\\
			\rowcolor{tabhighlight}
			CK-CoOp & \textBF{90.55} & \textBF{91.60} & \textBF{91.07}\\
			\bottomrule
		\end{tabular}
	\end{subtable}
	\\
	\begin{subtable}[t]{.3\textwidth}
		\centering
		\caption{FGVCAircraft.}
		\setlength\tabcolsep{2.5pt}
		\renewcommand{\arraystretch}{1.0}
		\begin{tabular}{l cc|c}
			\toprule
			& Base & Novel & H \\
			\midrule
			\gray{CLIP} & \gray{27.44} & \gray{35.77} & \gray{31.06} \\\midrule
			CoOp & \textBF{39.44} & 27.20 & 32.20 \\
			CoCoOp & 36.08 & \textBF{33.05} & \textBF{34.50} \\
			ProGrad & 39.43& 27.90& 32.68 \\
			\rowcolor{tabhighlight}
			CK-CoOp & 37.96 & 31.54 & 34.45\\
			\bottomrule
		\end{tabular}
	\end{subtable}
	\vspace{0.1em}
	\hspace{\fill}
	\begin{subtable}[t]{.3\textwidth}
		\centering
		\caption{SUN397.}
		\setlength\tabcolsep{2.5pt}
		\renewcommand{\arraystretch}{1.0}
		\begin{tabular}{l cc|c}
			\toprule
			& Base & Novel & H \\
			\midrule
			\gray{CLIP} & \gray{69.39} & \gray{75.53} & \gray{72.33} \\\midrule
			CoOp & {81.17} & 70.39 & 75.40 \\
			CoCoOp & 79.52 & \textBF{76.62} & {78.04} \\
			ProGrad & \textBF{81.36}& 73.95& 77.48\\
			\rowcolor{tabhighlight}
			CK-CoOp & 80.33 & 76.22 &\textBF{78.22}\\
			\bottomrule
		\end{tabular}
	\end{subtable}
	\vspace{0.1em}
	\hspace{\fill}
	\begin{subtable}[t]{.3\textwidth}
		\centering
		\caption{DTD.}
		\setlength\tabcolsep{2.5pt}
		\renewcommand{\arraystretch}{1.0}
		\begin{tabular}{l cc|c}
			\toprule
			& Base & Novel & H \\
			\midrule
			\gray{CLIP} & \gray{53.94} & \gray{57.97} & \gray{55.88} \\\midrule
			CoOp & {78.74} & 45.81 & 57.92 \\
			CoCoOp & 74.65 & 53.66 & {62.44} \\
			ProGrad & \textBF{77.31}& 48.95& 59.94\\
			\rowcolor{tabhighlight}
			CK-CoOp & 76.50 & \textBF{57.01} & \textBF{65.33}\\
			\bottomrule
		\end{tabular}
	\end{subtable}
	\\
	\begin{subtable}[t]{.3\textwidth}
		\centering
		\caption{EuroSAT.}
		\setlength\tabcolsep{2.5pt}
		\renewcommand{\arraystretch}{1.0}
		\begin{tabular}{l cc|c}
			\toprule
			& Base & Novel & H \\
			\midrule
			\gray{CLIP} & \gray{57.24} & \gray{64.08} & \gray{60.47} \\\midrule
			CoOp & {89.21} & 63.08 & 73.90 \\
			CoCoOp & 86.02 & 66.43 & {74.97} \\
			ProGrad & \textBF{90.76}& 65.25& 75.92\\
			\rowcolor{tabhighlight}
			CK-CoOp & 86.61 & \textBF{72.31} & \textBF{78.82}\\
			\bottomrule
		\end{tabular}
	\end{subtable}
	\hspace{30pt}
	\begin{subtable}[t]{.3\textwidth}
		\centering
		\caption{UCF101.}
		\setlength\tabcolsep{2.5pt}
		\renewcommand{\arraystretch}{1.0}
		\begin{tabular}{l cc|c}
			\toprule
			& Base & Novel & H \\
			\midrule
			\gray{CLIP} & \gray{71.04} & \gray{78.53} & \gray{74.60} \\\midrule
			CoOp & \textBF{84.35} & 65.21 & 73.56 \\
			CoCoOp & 82.28 & 70.87 & {76.15} \\
			ProGrad & 83.77& 68.36& 75.28\\
			\rowcolor{tabhighlight}
			CK-CoOp & 82.90 & \textBF{76.04} & \textBF{79.32}\\
			\bottomrule
		\end{tabular}
	\end{subtable}
\end{table*}

\subsection{Base-to-New Generalization}
\label{ssec:b2n}
The base-to-new generalization assesses the ability of a learned prompt context trained on a subset of classes to effectively generalize to a different task featuring a distinct set of classes. In this setting, for each dataset, the training phase employs the first half of the classes (\texttt{Base}), while the second half (\texttt{New}) is reserved for testing. It is important to note that the \texttt{Base} and \texttt{New} classes are mutually exclusive. We tune different learning-based methods on the training set composed of base classes and evaluate their performance by computing the accuracy on both base and new classes. Identical to previous studies~\cite{CoOp,cocoop}, the harmonic mean between the two accuracies is computed to provide a comprehensive measure of overall performance.

The summarized results with different backbones are reported in Table~\ref{tab:b2n_overall}. The average accuracies on base and new classes of all datasets are first computed, respectively, and their harmonic mean is further calculated. The absolute difference between the \texttt{Base} and \texttt{New} accuracies is shown in the table to reflect the generalization gap. The results obtained with different backbones exhibit similar trends, leading to the following observations.

\textbf{First}, among various prompt tuning methods, CK-CoOp significantly narrows the generalization gap compared to other methods, particularly CoOp and ProGrad. While CoCoOp also demonstrates a small generalization gap, its accuracies on both \texttt{Base} and \texttt{New} classes are notably lower than those achieved by CK-CoOp. \textbf{Second}, on average, CK-CoOp exhibits slightly lower \texttt{Base} accuracies compared to CoOp, with a maximum decline of 1.38\% in accuracy. This is expected as CK-CoOp introduces structural constraints on prompts, potentially limiting its representation ability to a certain extent. However, it is noteworthy that the accuracy reduction on base classes is relatively minor compared to the significant accuracy improvements on new classes, reaching a maximum of 6\%. In essence, the minor losses are outweighed by the substantial gains. Furthermore, CK-CoOp achieves the highest overall accuracy, measured by the harmonic mean, across all backbones. Specifically, the harmonic means are improved by 0.8\% to 1.1\% compared to the second best method. These findings confirm the superiority of CK-CoOp in base-to-new generalization.

Table~\ref{tab:base_to_novel}(a)-(k) list the detailed per-dataset results with ViT-B/16 as the backbone. As can be seen from the tables, CK-CoOp outperforms CoOp, CoCoOp and ProGrad on 11, 8 and 10 out of 11 datasets, under the harmonic mean metric. These results further confirm that the proposed structural context is more generalizable. When comparing CK-CoOp to CoOp and ProGrad, we observe relatively larger performance decreases in base classes for StanfordCars and FGVCAircraft. This is justifiable since we enforce a compact structure on the context, which may restrict its representation capabilities to some extent. Fine-grained tasks like StanfordCars and FGVCAircraft are more sensitive to such limitations. However, it is noteworthy that on these datasets, the accuracies on new classes are largely improved, effectively compensating for the aforementioned shortcomings.

\begin{table*}[t]
	\tabstyle{4pt}
	\caption{Results of domain generalization with different backbones using 16 shots.}
	\label{tab:dg}
	\renewcommand{\arraystretch}{1.0}
	\begin{tabular}{lcccccc}
		\toprule
		\multirow{2}{*}{Method}& Source & \multicolumn{4}{c}{Target} & \multirow{2}{*}{\textit{Average$\uparrow$}}\\
		\cmidrule(lr){2-2} \cmidrule(lr){3-6}
		& ImageNet & ImageNetV2 & ImageNet-Sketch & ImageNet-A & ImageNet-R & \\
		\midrule
		\multicolumn{5}{l}{\textbf{ResNet-101}}\\
		\gray{CLIP~\cite{CLIP}}&\gray{61.27}&\gray{54.84}&\gray{38.71}&\gray{28.33}&\gray{64.65}&\gray{49.56}\\
		CoOp~\cite{CoOp} & \underline{66.59} & 58.57 &39.50 & 29.07 & 63.30 &51.41\\
		CoCoOp~\cite{cocoop} & 65.35&58.43&\textBF{41.66}&\textBF{30.59}&\textBF{66.42}&\textBF{52.49}\\
		ProGrad~\cite{ProGrad}& \textBF{66.99} & \textBF{59.18} & 40.67 & \underline{30.01} & \underline{65.08} & \underline{52.39} \\
		\rowcolor{tabhighlight}
		CK-CoOp& 65.84 & \underline{58.82} & \underline{40.86} & 29.86 & \underline{65.08} & 52.09 \\
		\midrule
		\multicolumn{5}{l}{\textbf{ViT-B/32}}\\
		\gray{CLIP~\cite{CLIP}}&\gray{62.04}&\gray{54.79}&\gray{40.77}&\gray{29.51}&\gray{66.23}&\gray{50.67}\\
		CoOp~\cite{CoOp} & \underline{66.72} & 58.18 & 40.15 & 30.24 & 64.72 & 52.00\\
		CoCoOp~\cite{cocoop}& 66.07 & 58.19 & \textBF{42.23}& \textBF{32.13} & \textBF{67.00} & \textBF{53.12} \\
		ProGrad~\cite{ProGrad}& \textBF{66.98} & \textBF{58.73} & 41.53 & 31.04 & 66.03 &52.86\\
		\rowcolor{tabhighlight}
		CK-CoOp &66.36&\underline{58.31}&\underline{41.67}&\underline{31.95}&\underline{66.42}&\underline{52.94}\\
		\midrule
		\multicolumn{5}{l}{\textbf{ViT-B/16}}\\
		\gray{CLIP~\cite{CLIP}}&\gray{66.71}&\gray{60.89}&\gray{46.09}&\gray{47.77}&\gray{73.98}&\gray{59.09}\\
		CoOp~\cite{CoOp} & \underline{71.71} & 64.07 & 46.96 & 48.84 & 74.54 & 61.22 \\
		CoCoOp~\cite{cocoop} & 71.16 & 64.35 & \textBF{48.87} & \underline{50.73} & \underline{75.90} & 62.20\\
		ProGrad~\cite{ProGrad} & \textBF{72.19} & \textBF{64.82} & 48.48 & 49.96 & 75.81 & \underline{62.25}\\
		\rowcolor{tabhighlight}
		CK-CoOp & 71.40 & \underline{64.51} & \underline{48.55} & \textBF{51.10} & \textBF{76.10} & \textBF{62.33} \\
		\bottomrule
	\end{tabular}
\end{table*}

\begin{table*}[t]
	\tabstyle{6pt}
	\caption{Results under the cross-task generalization setting.}
	\label{tab:xd}
	\begin{tabular}{l c ccccccccccc}
		\toprule
		& Source & \multicolumn{10}{c}{Target} & \multirow{2}{*}{\rotbox{\emph{Average}\quad\quad}} \\ \cmidrule(lr){2-2} \cmidrule(lr){3-12}
		& \rotbox{ImageNet} & \rotbox{Caltech101} & \rotbox{OxfordPets} & \rotbox{StanfordCars} & \rotbox{Flowers102} & \rotbox{Food101} & \rotbox{FGVCAircraft} & \rotbox{SUN397} & \rotbox{DTD} & \rotbox{EuroSAT} & \rotbox{UCF101} &  \\
		\midrule
		\multicolumn{2}{l}{\textbf{ResNet-101}}&\\
		\gray{CLIP}&\gray{61.27}& \gray{85.85} & \gray{86.62} & \gray{63.19} & \gray{64.58} & \gray{80.39} & \gray{18.24} & \gray{59.05} & \gray{36.05} & \gray{32.59} & \gray{61.06} &\gray{58.99}\\
		CoOp & \underline{66.59}& 85.07 & 85.32 & 59.55 & 57.62 & 78.37 & 13.73 & 57.75 & 31.48 & 20.97 & 59.11 & 55.96\\
		CoCoOp & 65.35 & \underline{87.49} & \textBF{87.62} & \underline{61.84} & \textBF{63.59} & \textBF{81.41} & \underline{17.13} & \textBF{62.85} & \textBF{39.18} & \textBF{30.27} & \textBF{61.74} & \textBF{59.86}\\
		ProGrad & \textBF{66.98}& 85.94 & 86.62 & 59.49 & 58.36 & 80.33 & 15.73 & 60.90 & 34.18 & 26.17 & 59.20 &57.62 \\
		CK-CoOp & 65.83& \textBF{87.51} & \underline{86.89} & \textBF{61.87} & \underline{62.27} & \underline{80.68} & \textBF{17.59} & \underline{62.28} & \underline{37.71} & \underline{27.21} & \underline{60.33} & \underline{59.11}\\
		\midrule
		\multicolumn{2}{l}{\textbf{ViT-B/32}}&\\
		\gray{CLIP}&\gray{62.04}&\gray{90.10}&\gray{87.06}&\gray{60.15}&\gray{66.61}&\gray{80.90}&\gray{19.23}&\gray{62.05}&\gray{43.62}&\gray{45.52}&\gray{63.57}&\gray{61.90}\\
		CoOp&\underline{66.72}&88.23 & 85.52 & {57.72} & 59.58 & 79.25 & {15.27} & 59.34 & 35.30 & 31.49 & 59.58& 58.00\\
		CoCoOp&66.07 & \textBF{90.28} & \textBF{87.83} & \textBF{59.39} & \textBF{65.46} & \textBF{81.13} & \underline{17.73} & \textBF{64.68} & \textBF{42.12}& \underline{40.40} & \textBF{64.89} & \textBF{61.82}\\
		ProGrad&\textBF{66.98}&{88.84} & 86.36 & 57.32 & {60.56} & {80.31} & 15.21 & {63.05} & {38.59}& {37.08} & {61.77}&{59.64} \\
		CK-CoOp&66.36&\underline{89.13} & \underline{87.76} & \underline{58.67} & \underline{64.38} & \underline{80.95} & \textBF{17.80} & \underline{64.06} & \underline{38.93} & \textBF{41.32} & \underline{61.85}& \underline{61.02}\\
		\midrule
		\multicolumn{2}{l}{\textbf{ViT-B/16}}&\\
		\gray{CLIP}&\gray{66.71}&\gray{90.19}&\gray{88.90}&\gray{65.30}&\gray{70.19}&\gray{85.96}&\gray{24.85}&\gray{62.61}&\gray{44.21}&\gray{47.73}&\gray{66.75}&\gray{64.85}\\
		CoOp&\underline{71.71}&89.60 & 87.31 & 61.92 & 63.12 & 83.84 & 18.20 & 62.26 & 37.39 & 41.52 & 64.75 &61.97\\
		CoCoOp&71.16 & \underline{91.76} & \textBF{90.42} & \textBF{64.95} & \textBF{70.79} & \textBF{86.14} & \underline{22.57} & \textBF{67.34} & \textBF{45.45} & \underline{43.65} & \textBF{69.57} & \textBF{65.80}\\
		ProGrad&\textBF{72.24}&90.73 & 88.40 & 63.89 & 66.52 & 85.39 & 20.46 & 63.61 & 40.37 & 42.85 & 66.05 &63.68\\
		CK-CoOp&71.40&\textBF{91.80} & \underline{90.12} & \underline{64.32} & \underline{68.31} & \underline{86.12} & \textBF{23.20} & \underline{65.75} & \underline{42.97} & \textBF{48.08} & \underline{67.13} &\underline{65.38}\\
		\bottomrule
	\end{tabular}
\end{table*}

\subsection{Domain Generalization}
\label{ssec:dg}
This experiment studies the out-of-distribution generalization ability of CK-CoOp. In this setting, the context is optimized on source domain (i.e., ImageNet) and subsequently transferred to various target domains (i.e., ImageNetV2, ImageNet-Sketch, ImageNet-A and ImageNet-R) to assess the impact of distribution shift. The results are reported in Table~\ref{tab:dg}. We have the following observations: 1) CoOp demonstrates impressive performance on the source domain, but lags significantly on the target domains. This can be attributed to its excessive focus on the source domain. 2) CoCoOp significantly improves the accuracies on the target domains, especially on ImageNet-Sketch, ImageNet-A and ImageNet-R. However, this advancement comes at the cost of sacrificing accuracy on the source domain. 3) ProGrad excels at maintaining performance on the source domain, but falls short in generalization to ImageNet-Sketch, ImageNet-A and ImageNet-R. 4) CK-CoOp typically outperforms CoCoOp but lags behind ProGrad on ImageNet and ImageNetV2. However, on the remaining datasets, the situation is just the opposite. 5) In terms of the average performance, CK-CoOp achieves one first place and one second place, CoCoOp achieves two first places and ProGrad takes two second places. Overall, CK-CoOp and CoCoOp exhibit comparable domain generalization performance.

\begin{table}[t]
	\tabstyle{8pt}
	\caption{Multi-dimensional comparison from five dimensions with 16 shots. B2N: base-to-new generalization, DG: domain generalization, CT: cross-task generalization. Training and testing time are measured with a RTX A4000 GPU.}
	\label{tab:overall_compare}
	\renewcommand{\arraystretch}{1.0}
	\begin{tabular}{l  c c c  c c c}
		\toprule
		Method & \#Param$\downarrow$ & Training time$\downarrow$ & Testing time$\downarrow$ & B2N$\uparrow$ & DG$\uparrow$ & CT$\uparrow$ \\
		\midrule
		\textbf{ResNet-101}\\
		CoOp~\cite{CoOp} & \underline{8192} &\underline{1h 29min}&\textBF{1.76 ms}&69.22&51.41&55.96\\
		CoCoOp~\cite{cocoop} & 35360 &8h 53min&198 ms&70.86&\textBF{52.49}&\textBF{59.86}\\
		ProGrad~\cite{ProGrad} & \underline{8192} &2h 26min&1.88 ms&\underline{71.56}&\underline{52.39}&57.62\\
		CK-CoOp (ours) & \textBF{3106} &\textBF{1h 27min}&\underline{1.80 ms}&\textBF{72.36}&52.09&\underline{59.11}\\
		\midrule
		\textbf{ViT-B/32}\\
		CoOp~\cite{CoOp} & \underline{8192} &\underline{1h 27min}&\underline{1.92 ms}&70.05&52.00&58.00\\
		CoCoOp~\cite{cocoop} & 35360 &8h 48min&195 ms&71.12&\textBF{53.12}&\textBF{61.82}\\
		ProGrad~\cite{ProGrad} & \underline{8192} &2h 16min&1.96 ms&\underline{71.93}&52.86&59.64\\
		CK-CoOp (ours) & \textBF{3106} &\textBF{1h 24min}&\textBF{1.84 ms}&\textBF{72.74}&\underline{52.94}&\underline{61.02}\\
		\midrule
		\textbf{ViT-B/16}\\
		CoOp~\cite{CoOp} & \underline{8192} &\underline{1h 50min}& 2.04 ms&74.24&61.22&61.97\\
		CoCoOp~\cite{cocoop} & 35360 & 9h 21min &202 ms&\underline{76.04}&62.20&\textBF{65.80}\\
		ProGrad~\cite{ProGrad} & \underline{8192} & {2h 30min}&\underline{2.00 ms}&75.70&\underline{62.25}&63.68\\
		CK-CoOp (ours) & \textBF{3106} & \textBF{1h 49min}  &\textBF{1.96 ms}&\textBF{77.13}& \textBF{62.33}& \underline{65.38}\\
		\bottomrule
	\end{tabular}
	\vspace{-3pt}
\end{table}

\subsection{Cross-Task Generalization}
\label{ssec:cdt}
The cross-task generalization experiment aims to evaluate if the learned prompts on one dataset can be effectively transferred to other datasets. To be specific, the prompts are initially learned on ImageNet with 16 training images per class, and subsequently applied to classify the test images of 10 other datasets directly: Caltech101, OxfordPets, StanfordCars, Flowers102, Food101, FGVCAircraft, SUN397, DTD, EuroSAT and UCF101. The results are shown in Table.~\ref{tab:xd}. On the source task of ImageNet, ProGrad typically performs the best, even slightly better than CoOp. The results of CK-CoOp are slightly better than CoCoOp, however, both of them are slightly worse than CoOp. On the target tasks, CK-CoOp is outperformed by CoCoOp with a narrow margin. Notably, both of CK-CoOp and CoCoOp are significantly better than CoOp and ProGrad. On average, CK-CoOp can achieve comparable accuracy to CoCoOp across three different backbones. This confirms the good cross-task generalization ability of CK-CoOp.

\subsection{Multi-dimensional Comparison}
\label{ssec:muld_compare}
To conduct a thorough comparison of several methods, we summarize the prior generalization results in Table~\ref{tab:overall_compare}. Alongside these results, we also include information about the number of learnable parameters, training and inference speeds of various methods. Notably, the training and testing durations are obtained in the base-to-new setting, with training/testing sets of the base classes adopted. Additionally, the batch size for inference is 10. From the results, we have the following observations: 1) CK-CoOp outperforms CoCoOp in terms of base-to-new generalization accuracy across all backbones. It also demonstrates competitive domain and cross-task generalization performance in comparison to CoCoOp. It is noteworthy that CK-CoOp offers significant advantages in terms of lightweightness and efficiency, boasting a 91\% reduction in parameters, 80\% faster training speed and 100+ times faster inference speed compared to CoCoOp. 2) When compared to ProGrad, CK-CoOp exhibits a significant improvement in generalization performance, while simultaneously reducing the number of parameters by 62\% and saving $\sim$30\% in training time. In summary, the proposed CK-CoOp not only exhibits impressive generalization capabilities but also has excellent lightweightness and high efficiency.

\subsection{Ablation Study}
\label{ssec:ab_study}
This part validates the effectiveness of main components of CK-CoOp under three generalization settings. To inspect the impact of different configurations in detail, the results on base/new classes, source/target domain, source/target task are reported. As target domain/task contains several datasets, the target domain/task performance is obtained by averaging the accuracy across multiple datasets. Following previous works, the harmonic mean over base and new classes is reported for overall performance evaluation of base-to-new generalization. As for domain and task generalization, the mean of source and target domain/task accuracy is computed to reflect the overall performance. 

\textbf{Compositional Kronecker Structure.} To further validate the effectiveness of the compositional structure and the Kronecker product based bias, we conduct experiments by changing the context's structure with all other parameters fixed. The results are listed in Table~\ref{tab:effectness_ck}. The results from top to bottom in the table correspond to: 1) CoOp; 2) CK-CoOp without bias; 3) imposing compositional structure on the context of length $l$ directly; 4) the proposed CK-CoOp. 

By comparing row 2 to row 1, the overall accuracy exhibits a remarkable improvement, i.e., an accuracy gain of +1.60\%, +0.56\%, +1.81\% for base-to-new, domain and cross-task generalization, respectively. This validates our assumption that imposing compositional structure on the context is beneficial for generalization. By comparing row 3 to row 1, imposing Kronecker structure on context directly is not sufficient to achieve good results. We infer that the reason is that Kronecker structure limits the representation ability too much. By comparing row 4 and to 2, both the base and new accuracy are improved, implying the effectiveness of Kronecker product based bias matrix on boosting the representation ability of context. However, the bias has little effect on domain and task generalization performance, implying that there is essential difference between them and base-to-new generalization. The above results confirm that the proposed compositional Kronecker context is effective.

\begin{table}[t]
	\tabstyle{7pt}
	\caption{Effectiveness of compositional Kronecker structure. Comp: compositional structure, Kron: Kronecker product, B2N: base-to-new, DG: domain generalization, CT: cross-task generalization.}
	\label{tab:effectness_ck}
	\renewcommand{\arraystretch}{1.0}
	\centering
	\begin{tabular}{c c c  c c c  c c c  c c c}
		\toprule
		\multirow{2}{*}{Num.}&\multirow{2}{*}{Comp}&\multirow{2}{*}{Kron}&\multicolumn{3}{c}{B2N}&\multicolumn{3}{c}{DG}&\multicolumn{3}{c}{CT}\\
		\cmidrule(l{4pt}r{4pt}){4-6}\cmidrule(l{4pt}r{4pt}){7-9}\cmidrule(l{4pt}r{4pt}){10-12}
		& && Base & New & H & Source & Target & Avg. & Source & Target& Avg.\\
		\midrule
		1 & \ding{55} & \ding{55} & \textBF{82.29} & 67.63 & 74.24 & \textBF{71.71} & 58.60 & 65.15&\textBF{71.71}&\underline{60.89}&66.30\\
		2 & \ding{51} & \ding{55} & 80.91 & \underline{71.37} & \underline{75.84} & \underline{71.43} & \underline{59.98} & \underline{65.71} & \underline{71.43} & \textBF{64.78}&\textBF{68.11}\\
		3 & \ding{55} & \ding{51} & 66.39 & 67.23 & 66.81 & 63.04 & 55.13 & 59.09 & 63.04& 59.78&61.41\\
		4 & \ding{51} & \ding{51} & \underline{80.97} & \textBF{73.64} & \textBF{77.13}&71.40 & \textBF{60.06} & \textBF{65.73}&71.40&\textBF{64.78}&\underline{68.09}\\
		\bottomrule
	\end{tabular}
	\vspace{0pt}
\end{table}

\begin{table}[t]
	\tabstyle{8pt}
	\caption{Effect of sub-matrix size $s$.}
	\label{tab:submatrix_size}
	\renewcommand{\arraystretch}{1.0}
	\begin{tabular}{l c c c c c c c c c c}
		\toprule
		&\multirow{2}{*}{\#Param}&\multicolumn{3}{c}{B2N}&\multicolumn{3}{c}{DG}&\multicolumn{3}{c}{CT}\\
		\cmidrule(l{4pt}r{4pt}){3-5}\cmidrule(l{4pt}r{4pt}){6-8}\cmidrule(l{4pt}r{4pt}){9-11}
		&  & Base & New & H & Source & Target & Avg. & Source & Target& Avg.\\
		\midrule
		w/o bias & 0 & 80.91 & 71.37 & 75.84 & \underline{71.43} & 59.98 & 65.71 & \underline{71.43} & \underline{64.78}&\underline{68.11}\\
		$s=2$ & 35 & \underline{80.97} & \textBF{73.64} & \textBF{77.13}&71.40 & \underline{60.06} & \underline{65.73}&71.40&\underline{64.78}&68.09\\
		$s=4$ & 62 & \textBF{81.08}& \underline{72.14}& \underline{76.35} & \textBF{71.44} & \textBF{60.13} & \textBF{65.79} & \textBF{71.44} & \textBF{64.87} & \textBF{68.16}\\
		\bottomrule
	\end{tabular}
\end{table}

\begin{table}[t]
	\tabstyle{8pt}
	\caption{Effect of different normalization methods, different initialization methods of sub-matrices and quantization methods.}
	\label{tab:norm}
	\renewcommand{\arraystretch}{1.0}
	\begin{tabular}{l c c c c c c c c c}
		\toprule
		&\multicolumn{3}{c}{B2N}&\multicolumn{3}{c}{DG}&\multicolumn{3}{c}{CT}\\
		\cmidrule(l{4pt}r{4pt}){2-4}\cmidrule(l{4pt}r{4pt}){5-7}\cmidrule(l{4pt}r{4pt}){8-10}
		& Base & New & H & Source & Target & Avg. & Source & Target& Avg.\\
		\midrule
		\multicolumn{4}{l}{\textBF{Normalization methods}}\\
		w/o norm & 81.03 & 73.21 & 76.92 & \textBF{71.42}&59.96&65.69&\textBF{71.42}&64.45&67.94\\
		$L_1$ norm & 80.97 & \textBF{73.64}& \textBF{77.13}&\underline{71.40} & \underline{60.06} & \underline{65.73}&\underline{71.40}&\underline{64.78}&\underline{68.09}\\
		$L_1$ norm w/o first & \textBF{81.16} & \underline{73.29}& \underline{77.02}&\underline{71.40}&59.90&65.65&\underline{71.40}&64.28&67.84\\
		$L_2$ norm & 81.00 & 71.84& 76.15 & \textBF{71.42} & 60.04 & \underline{65.73} & \textBF{71.42} & \textBF{64.95} & \textBF{68.19}\\
		$L_2$ norm w/o first & \underline{81.12} & 72.99& 76.84 & \textBF{71.42} & \textBF{60.08} & \textBF{65.75} & \textBF{71.42} & 64.51 & 67.97\\
		\midrule
		\multicolumn{4}{l}{\textBF{Initialization of the sub-matrices}}\\
		zero with $L_1$ norm& \underline{3.90} & \underline{3.33} & \underline{3.59} & \underline{0.10} & \underline{0.50} & \underline{0.30} & \underline{0.10} & \underline{2.19} & \underline{1.15}\\
		normal & \textBF{80.97} & \textBF{73.64}& \textBF{77.13}& \textBF{71.40} & \textBF{60.06} & \textBF{65.73}&\textBF{71.40}&\textBF{64.78}&\textBF{68.09}\\
		\midrule
		\multicolumn{4}{l}{\textBF{Quantization methods}}\\
		Random Selection & \underline{80.92}& \underline{72.64}& \underline{76.56} & \textBF{71.41} & \underline{60.03} & \underline{65.72} & \textBF{71.41} & \underline{64.73} & \underline{68.07}\\
		K-Means & \textBF{80.97} & \textBF{73.64} & \textBF{77.13}&\underline{71.40} & \textBF{60.06} & \textBF{65.73}&\underline{71.40}&\textBF{64.78}&\textBF{68.09}\\
		\bottomrule
	\end{tabular}
\end{table}

\textbf{Sub-Matrix Size.} To investigate the impact of the sub-matrix size $s$ in the Kronecker product, we conduct a series of experiments by varying this parameter while maintaining all other parameters fixed. The results are listed in Table~\ref{tab:submatrix_size}. As evident from the table, increasing $s$ results in higher accuracy on base classes or source domain/task. This improvement is attributed to the enhanced representation ability enabled by the increased number of parameters. Furthermore, the strengthened representation ability initially benefits new classes or the target domain/task as well. However, it is worth noting that the optimal point where this benefit peaks varies depending on the type of generalization being evaluated. Specifically, when $s$ reaches or exceeds 4, the accuracy on new classes begins to decline. Nevertheless, it is noteworthy that even at $s=4$, there is no observable degradation in performance on either the target domain or task.

\begin{figure}[t!]
	\centering
	\hspace{-10pt}
	\begin{subfigure}{.6\columnwidth}
		\centering
		\includegraphics[width=1\columnwidth]{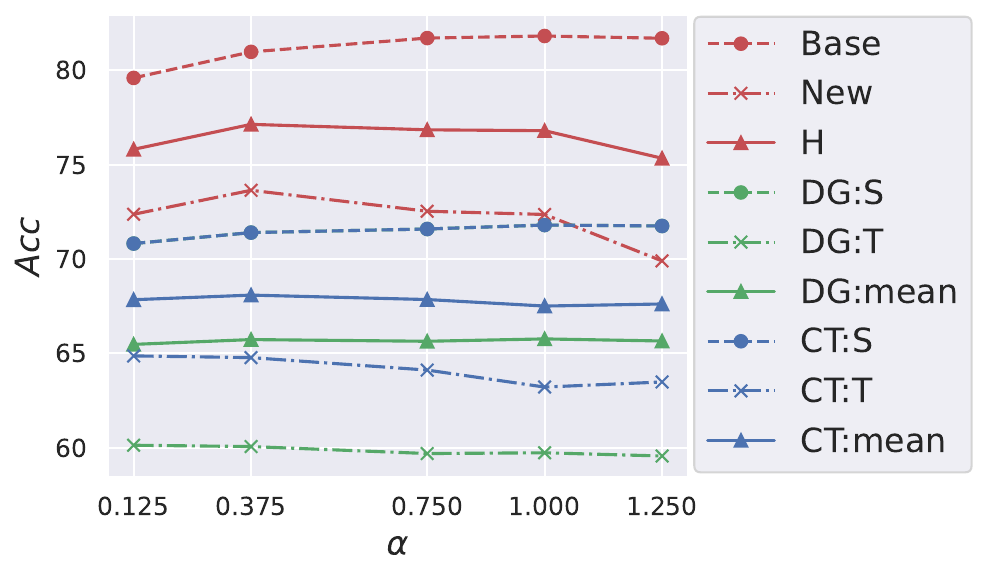}
		\caption{}
		\label{fig:sub1}
	\end{subfigure}%
	\hfill
	\begin{subfigure}{.39\columnwidth}
		\centering
		\includegraphics[width=1\columnwidth]{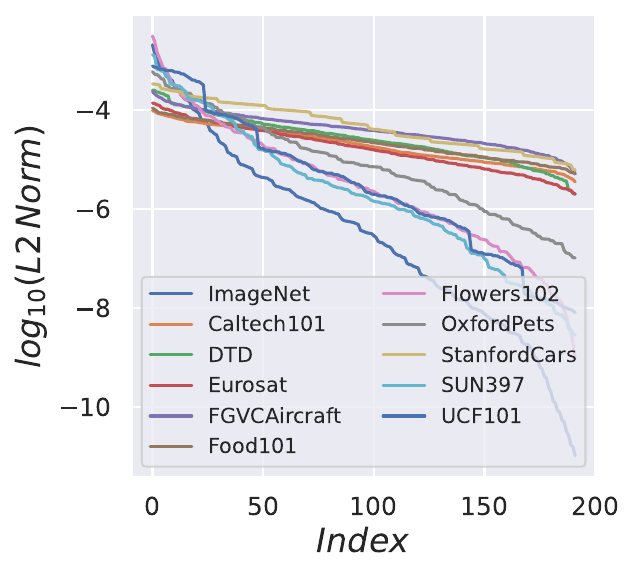}
		\caption{}
		\label{fig:sub2}
	\end{subfigure}
	\\
	\begin{subfigure}{1\columnwidth}
		\centering
		\includegraphics[width=0.98\columnwidth]{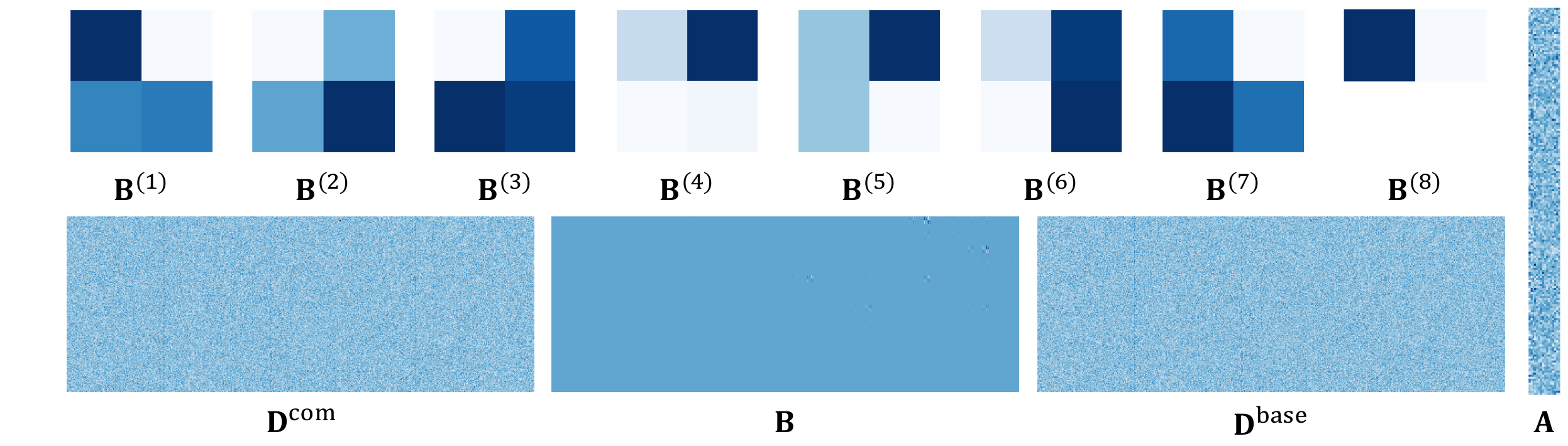}
		\caption{}
		\label{fig:sub3}
	\end{subfigure}%
	\caption{(a) Effect of $\alpha$ on base-to-new, domain and cross-task generalization performance. (b) Sorted $L_2$ norm of columns of bias matrix in logarithmic scale. (c) Visualization of various matrices in CK-CoOp. In (a), `DG:S', `DG:T', `DG:mean' denote domain generalization performance on source domain, domain generalization performance on target domain, and overall domain generalization performance. `CT:S', `CT:T', `CT:mean' denote cross-task generalization performance on source task, cross-task generalization performance on target task, and overall cross-task generalization performance.}
	\label{fig:CKCoOp_base2new_alpha}
	\vspace{0pt}
\end{figure}

\textbf{Normalization.} To validate the necessity of normalizing the sub-matrices, we compare the results with different normalization methods in Table~\ref{tab:norm}. The suffix ``w/o first'' denote the first sub-matrix $\mathbf{B}^{(1)}$ is not normalized. By doing so, the freedom of tuning the overall scale of the bias matrix is higher. Overall, the $L_1$ normalization works the best. It achieves +0.21\% and +0.98\% boost in harmonic mean accuracy compared to ``w/o norm'' and ``$L_2$ norm'', respectively. Meantime, its overall accuracy of domain generalization is comparable to ``w/o norm'' and ``$L_2$ norm'', and its cross-task generalization accuracy is better than ``w/o norm'' and slightly worse than ``$L_2$ norm''. Besides, we find that there is no distinct advantage to learn the overall scale.

\textbf{Initialization.} We conduct ablation study for different initialization methods of the sub-matrices. The zero (without normalization) and normal initialization (with $L_1$ normalization) are compared in Table~\ref{tab:norm}. The results show that the normal initialization can achieve significantly better performance. In fact, zero initialization without normalization does not work properly.

\textbf{Quantization Method.} Two methods to obtain the compressed dictionary are compared in Table~\ref{tab:norm}. `Random Selection' denotes some of pre-trained token embeddings are randomly selected to construct the compressed dictionary. The table shows that K-Means has a relatively big advantage over random selection for base-to-new generalization. However, the gains for domain and cross-task generalization are not that distinct. The advantage of K-Means lies in its optimization goal of minimizing quantization error, which reduces the information loss as much as possible.

\textbf{Dictionary Size.} The effect of the parameter $\alpha$ is studied in Fig.~\ref{fig:CKCoOp_base2new_alpha}(a). From the figure, the \texttt{Base} accuracy increases with larger $\alpha$. It nearly saturates as $\alpha$ approaches to 1.25. This can be attributed to stronger representation ability brought by larger space spanned by the base dictionary and more learnable parameters in the bias matrix and the coefficient matrix. The \texttt{New} accuracy first increases and then decreases with increasing $\alpha$. The underlying reason is the trade-off between representation and generalization ability. The overall accuracy reaches the maximal value when $\alpha$ approaches to 0.375. The influence of the parameter $\alpha$ on domain and cross-task generalization is not as strong. The accuracy of the source domain and source task slightly increases monotonically with the increase of $\alpha$. Significant accuracy degradation only occurs in the target domain and target task when $\alpha$ is greater than 1.

\begin{figure*}[t!]
	\centering
	\includegraphics[width=0.88\linewidth]{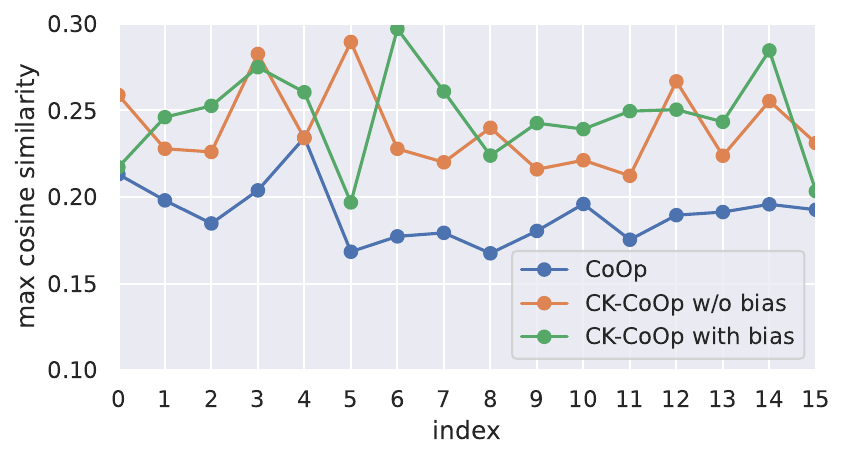}
	\caption{Statistics of max cosine similarities.}
	\label{fig:ctx_sim}
	\vspace{0pt}
\end{figure*}

\textbf{Visualization.} On each dataset, we compute the $L_2$ norm of the columns of the learned bias matrix. The values are plotted in Fig.~\ref{fig:CKCoOp_base2new_alpha}(b) in descending order for all datasets. As can be deduced from this figure, up to 25\% of the cluster centers in compressed dictionary undergo significant adjustments (with an $L_2$ norm larger than $10^{-4}$). This further confirms that the pre-trained token embeddings are already good candidates for building the prefix context. Besides, the amount of cluster centers that need tuning changes along with the used dataset. This can be attributed to different degrees of domain gap of class name text compared to upstream text data. Lastly, the matrices learned in CK-CoOp for the ImageNet dataset are visualized in Fig.~\ref{fig:CKCoOp_base2new_alpha}(c).

To further profile the underlying reason of better generalization, in Fig.~\ref{fig:ctx_sim}, we visualize the values of the max cosine similarity between each learned token embedding in context and all pe-trained token embeddings for CoOp, CK-CoOp without learnable bias and CK-CoOp with learnable bias. As CK-CoOp w/o bias learns context by directly combining pre-trained word embeddings, it could better maintain the pre-trained word space (reflected by higher similarities). The memory of such pre-trained knowledge is helpful for keeping the zero-shot generalization ability. CK-CoOp with bias has similar level of similarity but with stronger model capacity brought by bias. Although CoOp has adopted prompt template for context initialization, it can not maintain the pre-trained knowledge effectively, as reflected by the low similarities. In summary, the imposed structural constraint should be useful for improving generalization.

\section{Conclusion}
This paper introduces CK-CoOp, an innovative structural context optimization approach that significantly enhances the generalization capabilities of vision-language models (VLMs) when adapting to downstream tasks. CK-CoOp introduces a unique compositional context mechanism, which constructs context vectors by intelligently combining items from a dictionary composed of cluster centers derived from pre-trained token embeddings. This approach enables a more meaningful and interpretable representation of the contextual information, leading to improved generalization performance.

Furthermore, CK-CoOp incorporates a compact and learnable bias matrix, leveraging the Kronecker product, to further refine and enhance the representation capacity. This design not only maintains the expressiveness of the model but also achieves a better balance between efficiency and generalization.

Extensive experiments from multiple dimensions demonstrate the effectiveness of the proposed structured context optimization method. CK-CoOp exhibits stronger generalization ability while maintaining fewer parameters, shorter training time and efficient inference speed, making it a highly attractive choice for real-world applications.

Looking ahead, there are several potential directions for future work. One promising avenue is to explore other forms of structured methods to further enhance and generalize the contextual representation. This could involve investigating novel dictionary construction techniques or exploring alternative ways to combine items in the compositional context.
Another intriguing direction is to consider the co-design of the quantizer and compositional representation. By jointly optimizing these two components, we may be able to achieve even better generalization performance while maintaining the lightweight and efficient nature of CK-CoOp. This approach could lead to the development of even more powerful and versatile methods for tuning VLMs that are capable of handling a wider range of downstream tasks.

\bibliographystyle{unsrtnat}
\bibliography{main.bib}  

\end{document}